\documentclass[11pt,a4paper]{article}
\usepackage[hyperref]{emnlp2020}
\usepackage{times}
\usepackage{latexsym}

\usepackage{microtype}

\usepackage{url}
\usepackage{enumitem}
\usepackage{subfig}
\usepackage{booktabs}
\usepackage{makecell}
\usepackage{graphicx}
\usepackage{amsmath}
\usepackage{algorithm}
\usepackage[noend]{algpseudocode}
\usepackage{pgfplots}
\pgfplotsset{compat=1.14}
\graphicspath{ {./images/} }
\usepackage{enumitem}
\setlist[itemize]{noitemsep, topsep=0pt}
\setlist[enumerate]{noitemsep, topsep=0pt}
\usepackage{amsthm}
\usepackage{booktabs}
\theoremstyle{definition}

\usepackage{todonotes}
\usepackage[scaled=.7]{beramono}

\aclfinalcopy 
\def\aclpaperid{210} 

\title{Small but Mighty: New Benchmarks for Split and Rephrase}
\author{Li Zhang$^\spadesuit$\thanks{\hspace{3pt} Work done during internship at IBM Research.}\\
{}\\
\
\And
Huaiyu Zhu$^\diamondsuit$\\
{\hspace{100pt} $^\spadesuit$University of Pennsylvania \hspace{15pt} $^\diamondsuit$IBM Research \hspace{15pt} $^\clubsuit$ Google Research}\\
\texttt{\hspace{125pt} zharry@seas.upenn.edu \hspace{15pt} \{huaiyu,yunyaoli\}@us.ibm.com \hspace{15pt} sidbrahma@google.com}
\And
Siddhartha Brahma$^\clubsuit$\thanks{\hspace{3pt} Work done during employment at IBM Research.}\\
{}\\
\
\And
Yunyao Li$^\diamondsuit$\\
{}\\
}
\date{}

\begin{document}
\maketitle
\begin{abstract}
Split and Rephrase is a text simplification task of rewriting a complex sentence into simpler ones. As a relatively new task, it is paramount to ensure the soundness of its evaluation benchmark and metric. We find that the widely used benchmark dataset universally contains easily exploitable syntactic cues caused by its automatic generation process. Taking advantage of such cues, we show that even a simple rule-based model can perform on par with the state-of-the-art model. To remedy such limitations, we collect and release two crowdsourced benchmark datasets. We not only make sure that they contain significantly more diverse syntax, but also carefully control for their quality according to a well-defined set of criteria. While no satisfactory automatic metric exists, we apply fine-grained manual evaluation based on these criteria using crowdsourcing, showing that our datasets better represent the task and are significantly more challenging for the models.\footnote{Our datasets and code will be available at \url{https://github.com/System-T/TextSimplification}.}
\end{abstract}

\section{Introduction}
Split and Rephrase is the task of rewriting a presumably long and complex sentence into shorter and simpler sentences, while maintaining the same meaning. For example, one possible way to split the sentence ``Voiced by Aoi Koga, Kaguya is the series' titular character, popular among a wide audience.'' would result in ``Kaguya is voiced by Aoi Koga. Kaguya is the series' titular character. Kaguya is popular among a wide audience.'' While the split sentences have to be coherent, paraphrasing is not enforced. For example, the word ``titular'' does not have to be replaced. This type of text simplification is challenging as its natural language generation process potentially involves multiple sub-processes such as co-reference resolution, named-entity recognition, semantic role labelling, etc. Split and Rephrase has two main real-world uses: first, to benefit systems whose performance improves with decreasing length of sentences e.g. entity extraction \cite{zhang-etal-2017-position} and machine translation \cite{koehn-knowles-2017-six} by acting as a pre-processing step; second, to benefit human readers, especially those less proficient with the language when reading complex documents such as terms and agreements, in understanding the meaning more easily and accurately \cite{Inui2003TextSF,Siddharthan2002AnAF}. 

Datasets of the Split and Rephrase task contain pairs of a \textit{complex sentence} and a presumably meaning-preserving \textit{simplified rewrite} containing multiple simpler sentences. The task was introduced by \citet{narayan-etal-2017-split}, with the release of the WebSplit corpus. Afterwards, \citet{aharoni-goldberg-2018-split} proposed the state-of-the-art model to date, a sequence-to-sequence model \cite{Bahdanau2015NeuralMT} with a copy mechanism \cite{gu-etal-2016-incorporating,see-etal-2017-get} with the observation that most texts are unchanged during a Split and Rephrase operation. Later, \citet{botha-etal-2018-learning} introduced the WikiSplit corpus to be used as large but noisy training data, which the authors reported to be unsuitable as the evaluation data. Also, \citet{sulem-etal-2018-bleu} studied the problems of using BLEU as the evaluation metric for this task, while proposing a manually constructed test set called HSplit. 

We argue that the widely used benchmark dataset of Split and Rephrase, the WebSplit test set (known as simply WebSplit below), is not suitable for evaluation. Apart from its series of limitations already reported, such as a small vocabulary, unnatural expressions, etc. \cite{botha-etal-2018-learning}, we further show that its complex sentences systematically follow only 3 syntactical patterns marked by lexical cues (Section~\ref{sec:issues}). To demonstrate the implication of such limitations of WebSplit, we show that a simple, unsupervised rule-based model with only 3 corresponding operations can perform even slightly better than the state-of-the-art neural model (Section~\ref{sec:rule-basedModel}). 

To remedy the limitations of WebSplit, we crowdsource two new benchmarks with significantly more diverse syntax in the Wikipedia and legal contract domain with hundreds of human-written complex-simple sentence pairs (Section~\ref{sec:newBenchmark}). We carefully control for their quality based on 6 well-defined criteria of what constitutes a good Split and Rephrase rewrite. While most related work reports model performance using the widely criticized BLEU score and manual evaluation with no clear rubric, we perform fine-grained model evaluation using these 6 criteria, rated by crowd workers, showing that our benchmarks present models with greater challenges (Section~\ref{sec:modelPerformancee}).

\section{Issues with WebSplit}
\label{sec:issues}

WebSplit and Wiki-Split are two widely used datasets for the Split and Rephrase task. Because WikiSplit is derived from the edit history of Wikipedia, versions of passages are not necessarily written by Split and Rephrases operations, as the meaning may not be preserved during edits. Hence, WikiSplit is reported by its authors to be noisy and ill-suited for evaluation for this task \cite{botha-etal-2018-learning}. 

WebSplit is used in multiple previous works as the evaluation benchmark. It was created by automatically matching sentences in the WebNLG corpus \cite{gardent-etal-2017-creating} according to partitions of their meaning representations. The dataset has been shown to have various limitations, such as unnatural expressions, repetition of phrases \cite{botha-etal-2018-learning}, etc. 

Furthermore, our preliminary study shows that WebSplit contains several recurring syntactic patterns marked with lexical cues. To demonstrate this, we randomly sample 100 complex sentence from the test set, and are able to categorize them with only 3 syntactical patterns marked by lexical cues (underlined), at which some almost trivial Split and Rephrase operations can take place:\\
\textbf{relative clause (rc)} (48 out of 100): \textsf{\small
Scott Adsit voiced Baymax \underline{which} was created by Duncan Rouleau.}\\
\textbf{conjunction (conj)} (46 out of 100): \textsf{\small Above the Veil is from Australia \underline{and} was preceded by Aenir and Castle.}\\
\textbf{participle (part)} (13 out of 100): \textsf{\small \underline{Serving} the city of Alderney, the 1st runway is made from Poaceae.}

It can be further noticed that most complex sentences in WebSplit are short and require only one Split and Rephrase operation. We next show that a rule-based model which only exploits these patterns can perform on par with the state-of-the-art neural model on WebSplit. 

\section{Rule-Based Model}
\label{sec:rule-basedModel}
We design a simple rule-based model to exploit the syntactic cues widely present in WebSplit.

\subsection{Algorithm}

The rule-based model requires no training data and only uses semantic role labeling \cite{He2017DeepSR} and dependency parsing \cite{dozat2016deep}, running on AllenNLP \cite{gardent-etal-2017-creating}. Given a complex sentence, the model makes 3 splits when applicable. First, using semantic role labeling, the model identifies a Relational Argument and makes a split with the Relational Argument replaced by the Subject Argument. Second, The model looks for the word ``and'', making a split accordingly. Third, using dependency parsing, the model looks for a node which is joined by the clause, which is extracted, prepended with the subject, and split as a new simple sentence, while the rest of the original complex sentence is split as another new simple sentence.\footnote{The detailed algorithm is shown in the Appendix~\ref{appendix:rule}.}

\subsection{Performance}

The rule-based model and the state-of-the-art seq2seq model trained on WikiSplit \cite{aharoni-goldberg-2018-split,botha-etal-2018-learning} are evaluated using BLEU \cite{papineni2002bleu} on WebSplit. The rule-based model achieves a BLEU of 61.3, outperforming the neural model which achieves a BLEU of 56.0. The two models are also evaluated manually on 100 randomly sampled examples, with an identical accuracy of 64\% (the criteria of correctness is described in Section~\ref{phase2}). While the rule-based model is imperfect and can likely improve with more and better defined rules, it serves as a strong baseline that exploits the syntactical cues in WebSplit and potentially other benchmarks generated in a similar fashion. The strong performance of such a simplistic model highlights the need of more difficult and diverse benchmark data to better capture the complexity of the Split and Rephrase task.

\section{New Benchmark Datasets}
\label{sec:newBenchmark}
Considering the limitations of WebSplit, an ideal benchmark must not only be challenging with diverse patterns, but also ensure that the rewrites are strictly meaning-preserving Split and Rephrase. With these two goals, we collect two benchmark datasets, Wiki Benchmark (Wiki-BM) from Wikipedia and Contract Benchmark (Cont-BM) from the legal documents. These two datasets are to be used as gold standard for the evaluation of Split and Rephrase. To systematically control for the quality, we define 6 criteria of what constitutes a good Split and Rephrase, and validate the collected rewrites based on these criteria. 

\subsection{Collecting Complex Sentences}

First, we gather complex sentences as the input for the Split and Rephrase operation. 

\subsubsection{Wiki Benchmark (Wiki-BM)} While the simplified rewrites in the WikiSplit dataset are not guaranteed to be meaning preserving and cannot be used in a benchmark, the original complex sentences are semantically and syntactically diverse, with adequate complexity. From the 5000 complex sentences from the WikiSplit test set, we randomly select 500 for budget reasons with only alphanumerical characters, whites-spaces, commas and periods, and manually inspect them to ensure that they are well-formed. 

\subsubsection{Contract Benchmark (Cont-BM)} We collect sentences from publicly available legal procurement contracts online, and contract templates within IBM with no confidential information. We randomly sample and inspect 500 sentences in the same manners as above.

\subsubsection{Syntactical Diversity} To demonstrate that our complex sentences are syntactically diverse and are not plagued by patterns analyzed before, we randomly sample 100 complex sentences from each benchmark to annotate them by syntactical patterns. In addition to the 3 patterns outlined before, we define the following new patterns (the examples are truncated to save space):\\
\textbf{prepositional phrase (prep)}: \textsf{\small
The mausoleum was built in 1894 \underline{along} the lines specified by Frazer.}\\
\textbf{adverbial phrase (adv)}: \textsf{\small \underline{Except} as may be otherwise specified, Supplier shall invoice Buyer.}\\
\textbf{apposition clause (appos)}: \textsf{\small Leila married the movie director Ruy Guerra\underline{, }father of her only daughter.}\\
\textbf{infinitive clause (inf)}: \textsf{\small Nimfa was forced to take part of a devilish plan \underline{to} fool the Saavedra family.}

\begin{table}[t!]
\small
\centering
\begin{tabular}{ lccc } 
 \toprule
 Patterns & WebSplit & Wiki-BM & Cont-BM \\ 
 \midrule
 rc & 48 & 34 & 29\\ 
 conj & 46 & 71 & 66\\ 
 part & 13 & 34 & 28\\ 
 prep & 5 & 12 & 66\\ 
 adv & 0 & 19 & 38\\ 
 appos & 2 & 10 & 0\\ 
 inf & 0 & 5 & 10\\ 
 \midrule
 patterns/sent & 1.22 & 1.78 & 2.37\\ 
 \bottomrule
\end{tabular}
\caption{Counts of syntactic patterns for splitting in 100 random examples from each of WebSplit, Wiki Benchmark, and Contract Benchmark. Note that each complex sentence may have more than one pattern.}
\label{pattern-count}
\vspace{-1em}
\end{table}

\begin{table}[t!]
\small
\centering
\begin{tabular}{ lccc } 
 \toprule
 Entry & WebSplit & \makecell{Wiki-BM} & \makecell{Cont-BM} \\ 
 \midrule
 Rewritten by human & No & Yes & Yes\\ 
 \# complex & 930 & 403 & 406\\ 
 \# simple & 43958 & 720 & 659\\ 
 \# toks/complex & 20.6 & 29.6 & 41.5\\ 
 \# sents/simple & 3.7 & 3.0 & 3.0\\ \bottomrule
\end{tabular}
\caption{Comparison of statistics among WebSplit, Wiki Benchmark, and Contract Benchmark. }
\label{benchmark-comparison}
\vspace{-1em}
\end{table}

\begin{table*}[th]
\small
\centering
\begin{tabular}{ lcccccccc } 
 \toprule
    \textbf{WebSplit} & sensical & grammatical & \makecell{no miss\\fact} & \makecell{no new\\fact} & \makecell{correct\\split} & \makecell{enough\\split} & \textit{correct} & BLEU \\ 
    \midrule
    seq2seq & 71.6\%/4.55 & 64.0\%/4.40 & 94.30\% & 94.30\% & 87.30\% & 79.00\% & 50.1\% & 62.6 \\
    rule & 72.2\%/4.35 & 58.2\%/4.01 & 92.70\% & 95.70\% & 83.30\% & 82.90\% & 51.7\%  & 65.9 \\ 
    \toprule
    \textbf{Wiki-BM} & sensical & grammatical & \makecell{no miss\\fact} & \makecell{no new\\fact} & \makecell{correct\\split} & \makecell{enough\\split} & \textit{correct} & BLEU \\ 
    \midrule
    seq2seq & 55.4\%/4.22 & 47.8\%/3.93 & 98.30\% & 99.70\% & 80.30\% & 76.30\% & 37.3\%  & 87.0 \\
    rule & 59.8\%/4.06 & 54.2\%/3.85 & 94.30\% & 94.30\% & 78.70\% & 47.70\% & 28.5\%  & 77.2 \\
    human  & 84.9\%/4.76 & 76.8\%/4.61 & 95.00\% & 93.70\% & 88.30\% & 88.00\% & 68.4\%  & 77.8 \\ 
    \toprule
    \textbf{Cont-BM} & sensical & grammatical & \makecell{no miss\\fact} & \makecell{no new\\fact} & \makecell{correct\\split} & \makecell{enough\\split} & \textit{correct} & BLEU \\ 
    \midrule
    seq2seq & 29.4\%/3.45 & 25.0\%/3.04 & 92.70\% & 99.00\% & 52.30\% & 63.00\% & 16.7\%  & 78.6 \\
    rule & 57.9\%/3.97 & 54.8\%/3.89 & 97.70\% & 96.70\% & 83.30\% & 44.70\% & 25.0\%  & 79.2 \\
    human  & 78.2\%/4.53 & 72.6\%/4.43 & 95.30\% & 96.30\% & 93.70\% & 85.00\% & 63.3\%  & 73.0 \\
    \bottomrule

\end{tabular}
\caption{Average crowd ratings by criteria, model and benchmark. For the first two criteria which are on the scope of \textit{0}--\textit{5}, we report the percentage of \textit{5} and the average. For the rest which are \textit{yes}--\textit{no} questions, we report the percentage of \textit{yes}. }
\label{average-ratings}
\vspace{-1em}
\end{table*}

The counts from the manual annotation are shown in Table~\ref{pattern-count}. Wiki-BM has more diverse patterns and number of patterns per complex sentence than WebSplit, while Cont-BM has the most. The difference of complexity in the 3 benchmarks would be beneficial for evaluation. 

\subsection{Collecting Simplified Rewrites}

We ask a set of crowd workers to Split and Rephrase the gathered complex sentences on Amazon Mechanical Turk, and another set to ensure their quality\footnote{Detailed guidelines are shown in the Appendix~\ref{appendix:guidelines}.}. We divide the crowdsourcing workflow into two phases.

\subsubsection{Phase 1: Rewrite} For each complex sentence, we ask 3 crowd workers to rewrite it by splitting and rephrasing, with the option to flag the complex sentence as too simple or too problematic to split, which we later discard. We require Master Qualification, and pay \$0.2 per HIT for the complex sentences from Wiki-BM and \$0.4 per HIT for the more challenging Cont-BM. This Phase costs \$1,125 in total. 

\subsubsection{Phase 2: Rate} \label{phase2}
For each crowdsourced rewrite submitted in Phase 1, we ask 2 different crowd workers to evaluate its quality, based on the following fine-grained criteria:
\begin{enumerate}
    \item Is it sensical (scale of 0-5)?
    \item Is it grammatical (scale of 0-5)?
    \item Does it miss any existing facts (yes/no)?
    \item Does it introduce new facts (yes/no)?
    \item Does it have splits at the wrong place (yes/no)?
    \item Should some of its sentences be further split (yes/no)?
\end{enumerate}
We require Master Qualification, and pay \$0.07 per HIT\footnote{The pay exceeds the prorated US minimum wage.}. This Phase costs \$508. 

In each benchmark, we now have 500 complex sentences, each with 3 rewrites, each with 2 ratings. For each rating, if the worker answers 5 for the first two criteria, and chooses ``no'' for last four criteria, we denote this rating as \textit{correct}. For each rewrite, if both of its ratings are \textit{correct}, we denote this rewrite as \textit{perfect}. To ensure high quality of the gold standard, we only keep the rewrites that are \textit{perfect} as gold standard corresponding to their complex sentences in our benchmarks.

\subsection{Descriptive Statistics}

Some descriptive statistics and the comparison with WebSplit are shown in Table~\ref{benchmark-comparison}. While our new benchmarks are smaller than WebSplit, we argue that a small number of human-written, high quality ground-truth simple rewrites are better suited for evaluation than a larger number of automatically generated, noisy ones.

While similar to HSplit \cite{sulem-etal-2018-bleu}, our benchmarks include several additional features, such has  much more complex sentences from the legal domain, a clear set of rubrics for evaluation, and crowdsourced human judgements to scale.

\section{Model Performance}
\label{sec:modelPerformancee}
Previous work reports the model performance on this task using two metrics: BLEU on the entire benchmark and manual ratings on a small subset. However, BLEU has long been shown to have little correlation with human judgements in text simplification\footnote{We reinforce this claim in Appendix~\ref{appendix:correlation}.} \cite{sulem-etal-2018-bleu}. While other alternatives exist, the focus of our work is not the metrics, but rather the quality and difficulty of benchmarks, which can be illustrated no better than by human evaluation. Previously, manual evaluation has been done without a well-established rubric on what makes a Split and Rephrase rewrite correct. To address these problems, we use crowdsourcing following the process of Phase 2, by asking 3 crowd workers to rate model outputs based on the 6 fine-grained criteria described above.

Table~\ref{average-ratings} shows the average crowd ratings and BLEU score for each combination of a model and a benchmark. We consider the state-of-the-art seq2seq model trained on WikiSplit \cite{botha-etal-2018-learning} and our rule-based model. We use all rewrites from Phase 1 including those not included in our benchmarks to measure human performance. 

Both the rule-based and seq2seq model have large rooms for improvement, as they significantly underperform crowd workers in almost all criteria, with significantly lower performance in our proposed benchmarks than in WebSplit. Even for crowd workers, the percentage of overall \textit{correct} is less than 70\% in our new benchmarks, whose complex sentences are much more challenging to Split and Rephrase.

\section{Reliability of the Crowd}

Can we use crowdsourcing to evaluate models no less reliably as experts or authors, as done in previous work? As experts of this task, we manually rate a subset of model-output rewrites as the ground truth for rating, and compare it against the crowd's rating. Since there are 3 benchmarks and 3 models (including human, whose outputs are crowd rewrites we have collected in Wiki-BM and Cont-BM, \textbf{but not} WebSplit), there are 8 combinations in total. From the crowd ratings of these combinations, we assign each complex--output pair into one of 4 buckets, determined by the number of \textit{correct} ratings out of 3 crowd ratings. For each bucket, we sample 2 complex--output pairs. In total, $8\times 4\times 2=64$ complex--output pairs are sampled. The expert rates them independently following the same 6 criteria as the crowd workers. This gives the proportion of expert's \textit{correct} ratings among each bucket. 

\begin{figure}[t!]
\includegraphics[width=0.45\textwidth,keepaspectratio]{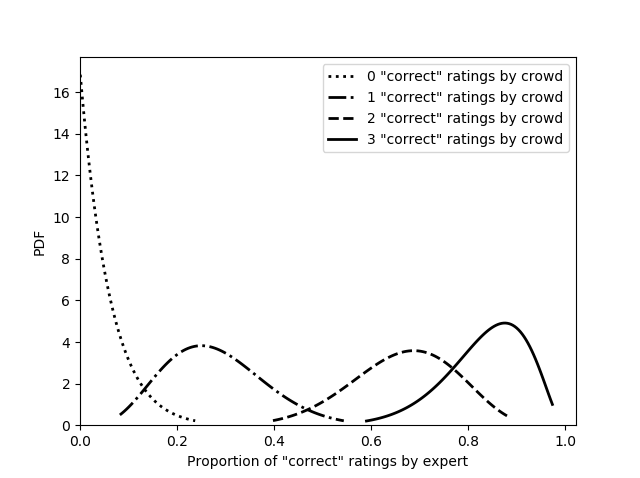}
\caption{Beta distributions with Laplace smoothing of the proportion of \textit{correct} ratings by expert. }
\label{crowd-reliability}
\vspace{-1em}
\end{figure}

These statistics allow us to fit a beta distribution for expert rating conditional on each crowd rating bucket, using Laplace prior smoothing.  The results are shown in Figure~\ref{crowd-reliability}. Each distribution corresponds to a bucket with 0, 1, 2, or 3 out of 3 \textit{correct} crowd ratings. For example, the right-most curve represents the probability density function where both the expert and the 3 crowd raters agree on a \textit{correct} rating. According to the figure with a 90\% one-sided confidence, when all 3 crowd raters rate a rewrite as \textit{correct}, the expert also rates \textit{correct} in more than around 80\% of the samples; when none of the 3 crowd raters rate a rewrite as \textit{correct}, the expert rates \textit{correct} for less than around 10\% of the samples.

This shows that crowdsourcing can be a reliable way to evaluate models for this task, with variable reliability depending on the number of raters per sample and their agreement. 

\section{Conclusion and Future Work}

After showing the flaws of the current benchmarks in Split and Rephrase, we release two crowdsourced benchmarks, Wiki Benchmark and Contract Benchmark, created from Wikipedia articles and legal documents respectively. Our benchmarks contain significantly more diverse syntax and provide additional challenges to models. Using fine-grained crowdsourcing evaluation on 6 well-defined criteria, we show that they provide a greater challenge to models.

We hope our benchmark datasets and human judgements facilitate model development and metric design, respectively. Moreover, future work should inspect the effect of Split and Rephrase on downstream tasks such as machine translation or information retrieval, and examine if models' performance on these tasks correlate with that on our benchmarks. 

\section*{Acknowledgments}
We thank Qing Lyu, Ranit Aharonov, Yannis Katsis, Lucian Popa, Nancy Xinru Wang and the anonymous reviewers for their valuable feedback. We also thank Shashi Narayan for providing helpful references on text simplification at the early stage of this work. 

\bibliographystyle{acl_natbib}
\bibliography{emnlp2020}

\appendix

\begin{table*}[t!]
\small
\centering
\begin{tabular}{ lcccccccc } 
 \toprule
    Benchmarks+Models& sensical & grammatical & \makecell{no miss fact} & \makecell{no new fact} & \makecell{correct split} & \makecell{enough split} & \textit{correct} \\ 
  \midrule
 WebSplit+s2s & .367 & .273 & -.037$\dagger$ & -.001$\dagger$ & .184 & .046$\dagger$ & .303\\ 
 WebSplit+rule & .491 & .456 & .118$\dagger$ & .048$\dagger$ & .425 & .276 & .480 \\ 
 Wiki-BM+s2s & .231 & .319 & .190 & -.005$\dagger$ & .167$\dagger$ & .256 & .412\\ 
 Wiki-BM+rule & .438 & .561 & .083$\dagger$ & .075$\dagger$ & .512 & -.035$\dagger$ & .232\\ 
 Cont-BM+s2s & .345 & .329 & .402 & -.062$\dagger$ & .191 & .215 & .255\\ 
 Cont-BM+rule & .277 & .190 & -.007$\dagger$ & .064$\dagger$ & .148$\dagger$ & .115$\dagger$ & .098$\dagger$ \\ 
  \midrule
 WebSplit+all models & .433 & .348 & .029$\dagger$ & .023$\dagger$ & .289 & .161 & .326 \\ 
 Wiki-BM+all models & .340 & .425 & .179 & .122$\dagger$ & .328 & .243 & .217 \\ 
 Cont-BM+all models & .313 & .271 & .228 & .01$\dagger$ & .199 & .142 & .165 \\   
  \midrule
 all benchmarks+s2s & .237 & .233 & .208 & .064$\dagger$ & .167 & .146 & .141 \\
 all benchmarks+rule & .388 & .400 & .081$\dagger$ & .063$\dagger$ & .347 & .089$\dagger$ & .230 \\   
  \midrule
 all benchmarks+all models & .362 & .393 & .172 & .068$\dagger$ & .315 & .168 & .251 \\   \bottomrule
\end{tabular}
\caption{Spearman's correlation between sentence-level BLEU and human judgement on 6 criteria by combinations of benchmarks and models. $\dagger$: the correlation coefficient is not statistically significant with $\alpha=.05$.}
\label{bleu-correlation}
\vspace{-1em}
\end{table*}

\section{Algorithm of the Rule-Based Model} \label{appendix:rule}
Given a complex sentence, the model runs the following processes once each.\\
\noindent\textbf{Wh Handling}
Using semantic role labeling, the model looks for a Relational Argument (R-ARG), and the Subject Argument (asserted to be the ARG preceding the R-ARG). Then, a split is made with the Relational Argument replaced by the  Subject Argument.\\ 
\textbf{Conjunction Handling}
The model looks for the word ``and''. Using semantic role labeling, if the word following ``and'' is an argument (ARG), assert that ``and'' is followed by a sentence, and a split is made. Or, if the word following ``and'' is a verb (V), the model asserts the Subject Argument to be the ARG preceding the V; a split is made with ``and'' replaced by the  Subject Argument.\\ 
\textbf{Insertion Handling}
Using dependency parsing, the model looks for a node with type \textit{participle modifier, relative clause modifier, prepositional modifier, adjective modifier,} or \textit{appositional modifier}. The clause with the node as the root is extracted,  prepended with the subject, and split as a new simple sentence. The rest of the original complex sentence is split as another new simple sentence. 

\section{Crowdsourcing Guidelines} \label{appendix:guidelines}

\subsection{Guidelines of Phase 1: Rewrite} 

\textbf{Instructions}: A long, complex sentence is hard to understand for many people. Please try to rewrite such a sentence by splitting and rephrasing it as several shorter and simpler sentences. A good example:
\begin{itemize}
    \item \textbf{Original}: Jonathan Thirkield, currently living in New York City, is an American poet who is known to be prolific.
    \item \textbf{Rewritten} (good): Jonathan Thirkield is an American poet. Jonathan Thirkield is known to be prolific. Jonathan Thirkield is currently living in New York City.
    \item \textbf{Rewritten} (good): Jonathan Thirkield is an American poet. He is currently living in New York City. He is known to be prolific.
\end{itemize}

Your rewrite must satisfy the following requirements:
\begin{enumerate}
    \item Grammatical\\
    \textbf{Rewritten} (bad: ungrammatical): Jonathan Thirkield currently \underline{living} in New York City. Jonathan Thirkield is an American poet. He is known to be prolific.
    \item Sensical and understandable \\
    \textbf{Rewritten} (bad: non-sensical): Jonathan Thirkield lives in \underline{prolific New York City}. He is an American poet.
    \item Has the same meaning as the original complex sentence, with no new facts and no missing facts (show/hide examples)\\
    \textbf{Rewritten} (bad: new facts): Jonathan Thirkield is a \underline{best-} \underline{selling} American poet. He is currently living in New York City. He is known to be prolific.\\
    \textbf{Rewritten} (bad: missing fact): Jonathan Thirkield is an American poet. He is currently living in New York City. (does not mention \underline{prolific})
    \item Split into appropriate number of short sentences (at least two), not too few or too many. If the sentence is too simple to be split, write SIMPLE as your response.\\
    \textbf{Rewritten} (bad: too few splits): Jonathan Thirkield, \underline{currently} \underline{living in New York City}, is an American poet. He is known to be prolific.\\
    \textbf{Rewritten} (bad: too many splits): Jonathan Thirkield is a poet. He is American. He is currently living somewhere. That somewhere is New York City. He is prolific. Such is known. (too many unnecessary splits)
    \item Do NOT use pronouns (it, she, he, they, this, that) if they are ambiguous\\
    \textbf{Rewritten} (bad: ambiguous pronoun): Walt Whitman is an American poet. Jonathan Thirkield is also an American poet. \underline{He} is living in New York City.
\end{enumerate}

Your rewrite will be validated by others. You might not receive payment if your rewrite does not satisfy the requirements. You may skip this HIT if you find splitting the given sentence too hard. However, if you manage to appropriately split a sentence which many other workers have skipped, you will receive a bonus.

\subsection{Guidelines of Phase 2: Rate}

\textbf{Instructions}: Read the two pieces of text below. The second text is an attempt to rewrite the first text, by splitting and rephrasing it into several shorter sentences to be understood more easily. Your job is to judge if this rewrite is good.

\begin{enumerate}
    \item The Rewritten text \textbf{makes sense}
    \item The Rewritten text \textbf{is grammatical}
    \item Does the Rewritten text miss some facts that are present in the Original text?
    \item Does the Rewritten text have new facts that are not present in the Original text?
    \item Does the Rewritten text split the Original text at the wrong place or unnecessarily?
    \item Does the Rewritten text have one or more sentences that should be further split?
\end{enumerate}

Each question is accompanied by a positive and negative example, the same as in the previous section. The crowd workers answer the first two questions by dragging a draw bar between ``Strongly Disagree'' and ``Strongly Agree'', and the last four questions by choosing ``yes/no'' radio boxes. 
\section{Correlation Between BLEU and Crowd Workers} \label{appendix:correlation}

Does BLEU correlate with human judgement on a large scale?
To answer this, we collect crowdsourced ratings of model outputs. With 3 benchmark datasets (WebSplit, Wiki-BM and Cont-BM) and two models (seq2seq and rule-based), we sample 100 complex sentence and output rewrite pairs from each combination, resulting in 600 in total.\footnote{Additionally, we sample 100 pairs each directly from Wiki-BM and Cont-BM with 3 crowd rewrites. These 600 pairs are used to measure human performance, but are not used in this section because they themselves are ground truth.} Then, we run the same crowdsourcing project as Phase 2 (Sec. 5.2.2) with these 600 pairs, for each of which we collect ratings from 3 crowd raters. The crowd raters are asked to rate based on the same 6 criteria as before (Sec. 3.1). As defined before, if a rating includes 5 for the first two criteria and ``no'' for the other four, it is considered \textit{correct}.

The Spearman's correlation coefficients between sentence-level BLEU and crowd ratings in each 6 criteria are shown in Table~\ref{bleu-correlation}. While BLEU has higher correlation with crowd raters on whether the rewrite is sensical or grammatical, most correlation coefficients are less than $.5$, and many do not imply a positive correlation at all. 

This reinforces the claim that BLEU is not a suitable evaluation metric for the Split and Rephrase task, because it has little correlation with human (crowd) judgement.

\end{document}